\documentclass[12pt]{article}

\usepackage{scicite}
\usepackage{amsmath}
\usepackage{subcaption}
\newcommand{\argminE}{\mathop{\mathrm{argmin}}}          % ASdeL
\usepackage{color}
\usepackage{float}
\usepackage{graphicx}
\usepackage{fourier} 
\usepackage{array}
\usepackage{makecell}
\usepackage{longtable}
\usepackage{graphicx}
\usepackage{arydshln}

\usepackage{multirow}
\usepackage{slashbox}
\usepackage[labelsep=endash]{caption}
\usepackage{hyperref}
\usepackage{times}

\newenvironment{sciabstract}{%
\begin{quote} \bf}
{\end{quote}}

\newcounter{lastnote}

\title{Derivative-free tree optimization for complex systems} 
\author
{Ye Wei$^{1,7*}$, Bo Peng$^{2,7}$, Ruiwen Xie$^{3,7}$, Yangtao Chen$^{4,7}$, \\ Yu Qin$^{5,7}$, Peng Wen$^{2,7}$, Stefan Bauer$^{4,7}$, Po-Yen Tung$^{6,7}$\\
\normalsize{$^{1}$School of Engineering, École Polytechnique Fédérale de Lausanne, Lausanne, Switzerland}\\
\normalsize{$^{2}$Department of Mechanical Engineering, Tsinghua University, Peking, China}\\
\normalsize{$^{3}$Institute of Materials Science, Technical University of Darmstadt, Darmstadt, Germany}\\
\normalsize{$^{4}$Helmholz AI,  M\"unchen, Germany}\\
\normalsize{$^{5}$School of Materials Science and Engineering, Peking University, Beijing, China}\\
\normalsize{$^{6}$Department of Materials Science and Metallurgy, University of Cambridge, Cambridge, UK}\\
\normalsize{$^{7}$Equal contribution}\\
\normalsize{$^\ast$Corresponding Author:
E-mail: ye.weinn@gmail.com}\\
\normalsize{}}
\date{}
\begin{document}
\baselineskip24pt
% Make the title.

\maketitle 
% Place your abstract within the special {sciabstract} environment.
% or performance of the surrogate model
%(e.g., local smoothness and convexity)
\begin{sciabstract}
%Don't stop me now!
A tremendous range of design tasks in materials, physics, and biology can be formulated as finding the optimum of an objective function depending on many parameters without knowing its closed-form expression or the derivative. Traditional derivative-free optimization techniques often rely on strong assumptions about objective functions, thereby failing at optimizing non-convex systems beyond 100 dimensions. 
Here, we present a tree search method for derivative-free optimization that enables accelerated optimal design of high-dimensional complex systems. Specifically, we introduce stochastic tree expansion, dynamic upper confidence bound, and short-range backpropagation mechanism to evade local optimum, iteratively approximating the global optimum using machine learning models. This development effectively confronts the dimensionally challenging problems, achieving convergence to global optima across various benchmark functions up to 2,000 dimensions, surpassing the existing methods 
by 10- to 20-fold. Our method demonstrates wide applicability to a wide range of real-world complex systems spanning materials, physics, and biology, considerably outperforming state-of-the-art algorithms. This enables efficient autonomous knowledge discovery and facilitates self-driving virtual laboratories. Although we focus on problems within the realm of natural science, the advancements in optimization techniques achieved herein are applicable to a broader spectrum of challenges across all quantitative disciplines.

\end{sciabstract}

\section*{Introduction}

%Materials design refers to a knowledge-guided approach to the development of high-performance materials. The strategy was established in the Bronze Age and has undergone further developments since that time. Materials design is the basis for the scientific development that enables technological progress. Several thousand different material systems have been developed so far that serve in engineering applications. The conventional deterministic paradigm in materials design, rooted in the assumption of fixed and precise material properties, has encountered limitations in addressing the inherent variability present in real-world scenarios. The emerging field of Probabilistic Design of Materials (PDM) represents a pivotal shift toward a more comprehensive and nuanced approach in engineering and materials science.

%Simulation of the process of arriving at an optimal design by annealing under the control of a schedule is an example of an evolutionary process modeled accurately by purely stochastic means. It may be a better model of selection processes in na
%A tremendous range of design problems in  
Derivative-free optimization (DFO), which refers to searching for global optima without knowing the structure, gradient, and convexity of the objective function, is ubiquitous. A wide variety of important areas of science and engineering came under the dominion of the theory of DFO. For example, in materials science, compositionally complex alloys (CCAs) design can be reformulated as the DFO of the properties by tuning the alloy composition \cite{Zhang2012}; In biology, the protein design can be approached as a DFO problem of bio-functionalities by optimizing the amino acid sequence \cite{RICHARDSON1989,Bassil1997}. 

The history of optimization can be traced back to 17-18th century, when Leonhard  Euler and Joseph-Louis Lagrange developed methods for finding maxima and minima of functions of several variables subject to constraints, while Newton and Gauss pioneered iterative methods to approach an optimal solution, laying the groundwork for what would later be known as calculus of variations, a central instrument to many quantitative disciplines such as the analytic mechanics, geodesics and variational methods \cite{bruce2004,kimmel1998}. 
A more modern conception of optimization originates in the seminal works of George Dantzig's work on linear programming in the 1950s \cite{Bertsekas2005}. Karush, Kuhn and Tuck developed the Karush–Kuhn–Tucker conditions, providing the necessary conditions for optimality in nonlinear programming problems \cite{Boyd2004}. 
Inspired by natural processes, evolutionary and heuristic approaches were developed in the late 20th Century, which have become standard toolkits for engineering and scientific communities to find approximate solutions in which acquiring an exact solution is impractical\cite{Hansen2001, Storn1997,TSALLIS1996395}. The success of such methods was evidenced by the discovery of near-optimal solutions in a wide range of NP-hard (non-deterministic polynomial-time hard) problems, such as the traveling salesman problem and integrated circuit design \cite{Karp2010, Kirkpatrick1983}. Recently, optimization techniques have found new applications in artificial intelligence. In particular, gradient-based optimization techniques have become instrumental in training deep learning models, giving rise to the era of deep learning\cite{wright2012}.

%with the advent of advanced computing and machine learning
The essential prerequisite for DFO is an objective function that quantitatively measures the "goodness" of a solution given the set of variables. The gradient-based optimization requires a differentiable objective.
However, it is well accepted that the knowledge of the internal workings of many real-world systems is not fully accessible and the derivative regarding the objective function is unknown \cite{Brockhoff2016}. We refer to these types of non-convex, non-differential systems as 'complex systems'. One way to address this problem is to treat the complex system as a 'black-box' and learn a surrogate model to approximate the objective function and optimize the design through the learned model\cite{andrew2009}.

Nevertheless, the true nature of a complex system is complicated and diverse: designing a complex system often means traversing a vast solution space characterized by dozens to hundreds of variables. Both exact and heuristic approaches struggle to adapt and scale to such high-dimensional scenarios, often exhibiting reduced robustness and leading to suboptimal results. Another possible way for DFO is the learning-based approach, many previous works, such as Bayesian Optimization (BO) and its variants \cite{Shahriari,Bubeck2011,Springenberg2016} learn a Bayesian model of the objective function and sample the best candidates using uncertainty-based technique such as Thompson sampling\cite{Bobak2018}, which works well when the objective function is low-dimensional (e.g., below 20 dimensions) and can be well approximated but often fails at higher dimensions \cite{Frazier2018}. Recent works explore tree search method, which is the key component of many revolutionary AI algorithms such as AlphaGo\cite{Silver2016}. The tree search methods generally rely on upper confidence bound ($UCB$) or its variants to achieve the optimal balance between exploration and exploitation \cite{Auer20021, Auer20022}. These methods partition the search space iteratively \cite{Kim2020} and use local surrogate models to approximate the promising search subspace\cite{Eriksson2019}. However, their performance depends on the local models and also cannot handle the curse of dimensionality \cite{Wang2020}. Worse still,  nonlinear phenomena are pervasive in complex systems. The highly nonlinear objective space characterized by myriad local minima could lead to the search algorithm being trapped locally for a very long time and struggling to converge to global optima.
% A typical example is the critical point at the phase transitions,  whose certain thermodynamic properties could diverge and violate the Lipschitz continuity. Local minima, characterized by drastic changes in objective function between neighboring states, bear some similarities to the critical points and are intrinsically difficult to model. 
%expected improment
%, easily resulting in memory explosion
%the critical-point-like local minima could strongly influence the search due to their singular behavior, resulting in non-physical and erroneous search statistics.
%So far, there has been little progress in this regard.
% information about the derivative of the objective function f is unavailable, unreliable or impractical to obtain.
% local-iterative%Ever since its112 proposal by Kocsis and Szepesvári in the early 21st century [], it has been the key component113 of many revolutionary AI algorithms such as AlphaGo

%To avoid the accumulation of potential non-physical statistics induced by divergence, 
We introduce a Derivative-free stOchastic Tree Search (DOTS) method for the optimization of high-dimensional complex systems. We do so by constructing a stochastic search tree with a short-range backpropagation mechanism and a dynamic upper confidence bound ($DUCB$). The new backpropagation updates the search information locally, while the new upper bound incorporates an adaptive exploration term, dynamically adjusting the balance between exploration and exploitation based on the data distribution. Additionally, DOTS introduces a novel sampling technique, sampling both top-scored and most-visited states, thereby reducing the number of data points required to attain the global optimum.
%by training a deep neural network as a surrogate objective function and 
% A more informative exploration-exploitation strategy should account for the performance of previous rounds (
%, which enables efficient escape of the local minima

We conduct a thorough benchmarking of DOTS against various state-of-the-art (SOTA) algorithms across six types of test functions. Remarkably, The DOTS outperforms all SOTA methods by an order-of-magnitude, achieving 100$\%$ convergence ratio on major test functions (excluding Rosenbrock, which is 80$\%$) with dimension ranging from 20 to 2,000 with as few as c.a. 500 data points. Further, we present a comprehensive numerical analysis and ablation study to elucidate the contribution of individual components. Finally, we integrate DOTS, machine learning models and high-fidelity simulators to construct self-driving virtual laboratories (SVL) for real-world complex systems. In these applications, DOTS exhibits superior performance compared to existing methods, demonstrating its wide applicability and high versatility. Through detailed investigations into the underlying mechanisms of DOTS designs, we validate its ability to navigate vast search spaces and autonomously discover new knowledge across different disciplines without requiring human intervention.
\section*{Results}

\subsection*{Problem statement}
In the DFO, we have a function $f$ without explicit formulation and the goal is to find a specific state $x^*$ with the best scores. Without loss of generality, assuming that we search for global minima:
\begin{equation}
x^* =  \argminE_{x \in X} f(x)
\end{equation}
where $x$ is the input vector and $X$ is defined as the search space, typically $R^N$, and N is the dimension. $f$ is the deterministic function that maps the input $x$ to the label, which can either be an exact function that provides ground-truth labels or a data-driven surrogate model learned through the dataset $D = \{{(x_i, y_i)}\}^N_i$, in which $N$ is the number of labels and $y_i$ is the label of $x_i$. It is noteworthy that this function is not limited to single-objective problems, it can be a product of multiple functions as long as it solely depends on $x$, which makes it a multi-objective task\cite{WU201990}. 
In addition,  unlabeled data is often abundant in real-world tasks, but the label is rather scarce and expensive to obtain. One common solution is active learning, which approaches $x^*$ by iteratively searching and sampling from a surrogate model $f$. Finding the global optima $x^*$  of the real-world complex system requires three essential components: 1) A parameterized form of the complex system; 2) Search and sampling; 3) A well-trained surrogate model. The benchmark study shows that the neural network outperforms other learning models (Extended Data Fig. 1). Thus, we fixate on the backbone of neural network and focus on the first two components throughout the rest of the work (see Methods and Supplementary Notes for the technical details of the machine learning models).
%the model $f$ and the sampling strategy significantly affect the convergence ratio.  
% derivtative-free/black-box optimization by tree search
\subsection*{Feature engineering of complex systems}
A fundamental requirement for optimizing the complex system is proper feature engineering, such that one can use a numerical vector or matrix to accurately represent the state of the complex system;  The node in the tree search is defined as the feature vector of the system obtained by feature engineering. To handle real-world complex systems, the input 
$x$ is discretized, as the feature vectors often possess discrete or limited resolution.
For instance, in protein design, the protein sequence can be represented by the 20 natural amino acids, which can be encoded into discrete integer numbers ranging from 1 to 20 (Fig. 1a). Alternatively, one can approximate continuous systems by discretizing features into smaller intervals. While DOTS may require more actions to reach the global optimum with smaller intervals, the convergence behavior of the DOTS algorithm is generally unaffected by discretization (Supplementary Fig. 1).

%The degree of discretization is rather problem-specific,

In the DOTS algorithm, the root node initiates the generation of leaf nodes (Fig. 1b). This process involves taking actions defined as numerical changes to the feature vector. There are three modes of action, each occurring with equal probability (1/3): 1) One-step move: This mode represents the smallest possible change at a single position of the feature vector. For example, if the smallest possible interval is 0.1, then the smallest possible change is $\pm$0.1; 2) Single mutation: In this mode, one position of the feature vector randomly mutates to any value within the allowed range; 3) Scaled random mutation: This mode involves a proportion of the feature vector randomly mutate to any allowed values.
The number of leaf nodes equals the dimension of the feature vector (Fig. 1c). This process of generating leaf nodes is termed "stochastic expansion".
%The center idea in the back of MCTS is to build a seek tree incrementally by using simulating more than one random performs (regularly known as rollouts or playouts) from the current recreation nation. 
%The focus of MCTS is on the analysis of the most promising moves, expanding the search tree based on random sampling of the search space. The application of Monte Carlo tree search in games is based on many playouts, also called roll-outs. In each playout, the game is played out to the very end by selecting moves at random.
%in the alloy design, the alloy system can be represented using a numerical vector, representing the content of each metallic element in terms of percentage. 
\subsection*{Derivative-free stochastic tree search}
DOTS consists of two basic components: conditional selection and stochastic rollout. It performs the optimization search by repeatedly conducting the conditional selection and stochastic rollout till the stopping criteria are met (e.g., 100 rollouts). We summarize the essential features of DOTS (shown in Fig. 1c-d) as follows: 1) $DUCB$, which is a modified UCB formula for dynamically balancing the exploration and exploitation trade-off (Fig. 1e). 2) The conditional selection, which is based on an inequality of $DUCB$, namely, the stochastic rollout proceeds with the leaf node that has lower $DUCB$ than the root node, otherwise it proceeds with the same root;   
3) Stochastic rollout, which consists of two components. The first component is the scaled stochastic expansion of the root nodes and the second one is the local backpropagation designed to escape from local minima by iteratively updating the visitation information between the root node and the selected leaf node. 
We provide a detailed comparison between classic Monte Carlo tree search and DOTS in Supplementary Notes and Supplementary Fig. 2.
%is one of the key ingredients of DOTS algorithm,
%Additionally it does not update any earlier nodes along the search trajectory. Also, it does not backpropagate the value information, so that the values of nodes do not affect each other.
% repeatedly conducting the conditional selection and stochastic rollout (Fig. xx).
% DOTS adopts a similar workflow as MCTS (both contain selection and rollout steps), but significantly differs in terms of e.
%Conventional backpropagation step updates both the value and visitation information of all nodes along the path,  whereas 

%Finally, we showcase an active learning loop that combines the DOTS, machine learning surrogate model, and a search-statistics-based sampling technique that could achieve state-of-the-art results not only on various synthetic test functions but also optimization problems in a wide range of scientific domains (Fig. 1G).
%in the following discussion, we will discuss their difference in detail.
\begin{figure}[H]
    \centering
    \includegraphics[width=1\textwidth]{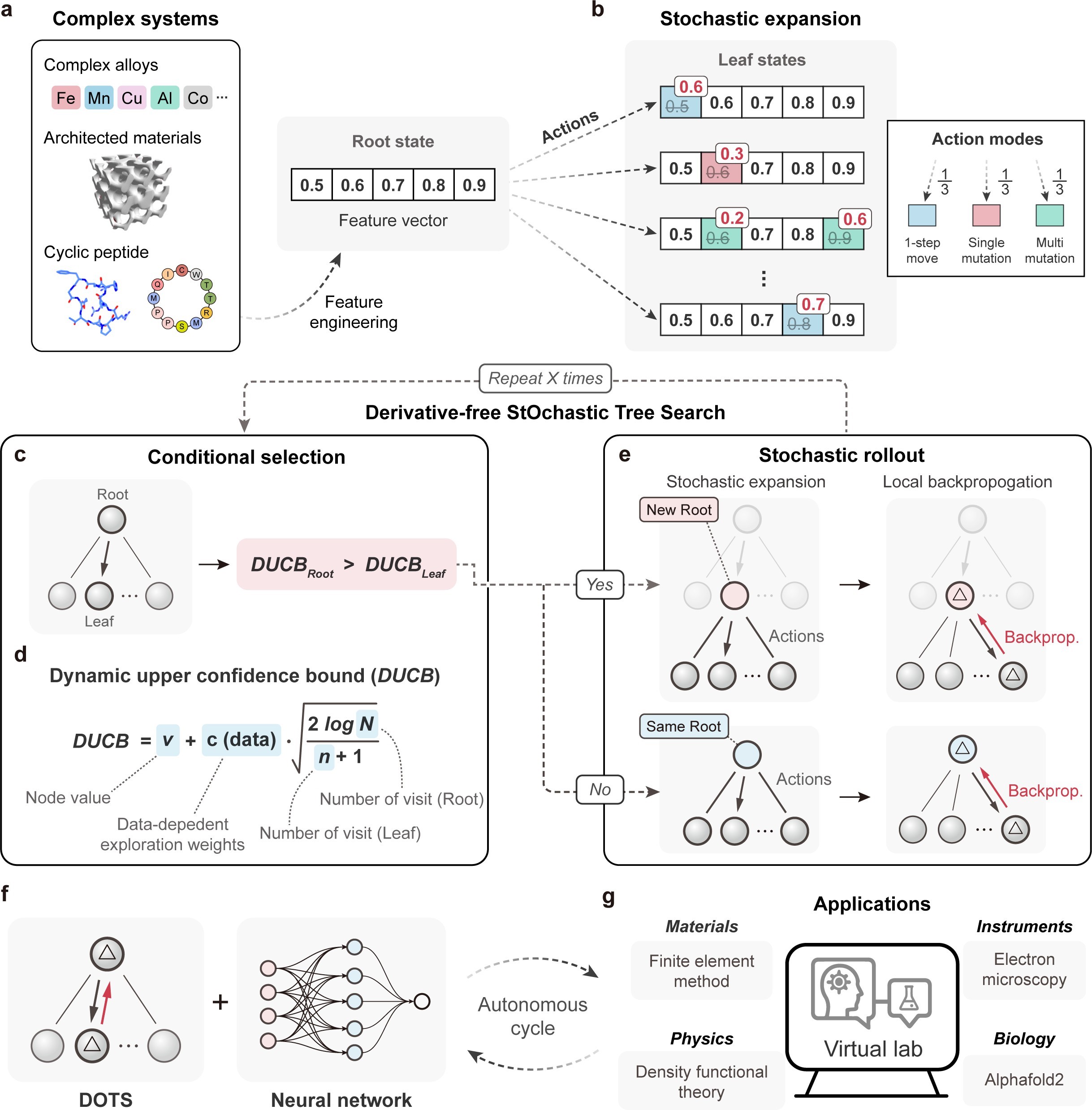}
    \caption{\textbf{Principle of derivative-free stochastic tree search.} \textbf{(a)} Feature engineering of complex systems. \textbf{(b)} Stochastic Expansion. \textbf{(c)} Conditional selection. \textbf{(d)} Stochastic rollout. \textbf{(e)} Dynamic confidence upper bound. \textbf{(f and g)} An autonomous virtual lab consisting of DOTS, neural network and simulators designed for various DFO applications in materials, instruments, physics, and biology. 
    }
    \label{fig:1}
\end{figure}
%In the following discussion, we will describe each step and their differences in detail.
\subsection*{Conditional selection}
The key component of classic tree search for balancing the exploration and exploitation is $UCB$, which can be expressed as follows: 
\begin{align}
    UCB &= v_{average} + c\cdot \sqrt{\dfrac{2logN}{n}}
\end{align}
$ v_{average}$ is the average of the sum value of all nodes along the path (from the starting to the end node). $c$ is the exploration weight, $N$ is the total number of visitation (number of visits of the root node) and $n$ is the visit number of the current node. Due to the high dimensionality of the complex system, the vast majority of the node is unvisited, so the $n$ equals zero and the corresponding $UCB$ is infinite. Consequently, the classic tree search needs to visit all child nodes at least once to perform select and rollout steps before selecting the node with finite $UCB$, resulting in an extraordinarily heavy computational burden. To alleviate this burden, $DUCB$ adds 1 to the denominator of the original $UCB$ formula, equivalent to assuming that all nodes have been visited once. Hence, $DUCB$ always yields a finite value with which all leaf nodes can be compared, without the need to perform select and rollout steps for each individual, greatly accelerating the search process. Moreover, the $DUCB$ is data-dependent, dynamically changing the value of exploration weight based on the ground-truth distribution. The Dynamic $UCB$ formula is written as follows:
\begin{align}
%UCB &= \hat{v} + c\cdot \sqrt{\dfrac{2logN}{n}}\\
DUCB  &= v_{node} + c(\rho)\cdot \sqrt{\dfrac{2logN}{n+1}}\\
c(\rho) &= c_0\cdot min(\rho)
\end{align}
where $v$ represents the value of the current node. With this modification, we can directly compare the $DUCB$ of all child nodes simultaneously and select the one with the highest  $DUCB$. $\rho$ is defined as the label distribution and $c_0$ is a constant, which is a hyperparameter (between 0.1 to 1).
The conditional selection is based on the $DUCB$ inequality. During this step, if the $DUCB$ of the root node is lower than that of all leaf nodes, then DOTS proceeds with the same root node in the next round, otherwise, the leaf node with lower $DUCB$ becomes the new root and proceeds to the stochastic rollout.
In the classic setting of tree search, $UCB$ has a fixed exploration weight $c$, whereas $DUCB$ adopts an adaptive exploration weight $c(\rho)$ (Fig. 2j): the weight $c$ is no longer a fixed constant but proportional to the minimum value of distribution $\rho$ (Equation 4). This is because in the DFO task, the newly sampled data could significantly affect the label distribution. Thus, the exploration weight should scale with the new minimum to keep the balance between exploitation and exploration.
%\subsection*{}
%in which the statistics would sequentially accumulate from the leaf all the way to the root nodes (long-range)

%The exploration weight should depend on the previous round of performance, i.e., it should increase at the current round if earlier iterations show little progress. More details can be found in the Methods. 
% and the history of progress $h$.
%The DOTS adopts a dynamic exploration-exploitation strategy by adjusting the exploration weight based on the previous round of performance and label distribution. Specifically, the exploration weight should increase drastically at the current round if earlier iterations show little progress.  More details can be found in the Methods.
%mediocre performance of the earlier rounds should be compensated by more aggressive exploration at the current round

\subsection*{Local backpropagation}
%Stochastic rollout consists of two building blocks: stochastic expansion and local backpropagation.
The classical backpropagation mechanism uses the result of the rollout to update both value and visitation of the nodes along the path, which affects all nodes (from root to end node) at a global level. The local backpropagation only updates the visitation information of the current root node and the subsequent leaf nodes (Fig. 1d). We do not update the value information, since our optimization problem only concerns the discovery of a single optimal state and has little 'memory' of previous states. Such modification has two major effects: 1) The value information is not backpropagated and the visitation backpropagation is short-ranged, preventing the information of local minima from backpropagating along the search trajectory;  2) The distant information is completely forgotten, and DOTS only utilizes nearby information to perform exploration.
%3) It creates a local 'artificial gradient' that enables the DOTS to jump out of the local optima.

\subsection*{Top-visit sampling}
Sampling technique becomes critical when a machine learning surrogate model is used. An ideal sampling method should choose the most promising data while keeping the diversity of the data distribution (both input and label) to ensure that the surrogate model can generalize to unseen data. The most straightforward sampling approach for tree search is to sample the data with the highest predicted scores. However, the distribution of the top-scored samples from a flat objective space is rather narrow, which could render a rather homogenous data distribution and lead to the overfitting of the surrogate model. One way to increase the generalization power of the surrogate model is to sample the most-visited nodes, despite their lower scores. In other words, they are some local minima (other than top-scored nodes) that trap the algorithm for a while, and learning from the distribution of these local basins could help the surrogate model understand the underlying landscape of objective function. Based on this argument, we introduce a new sampling technique and coin it 'Top-visit sampling'(Fig. 2g). Top-visit technique samples both the top-scored and the top-visited nodes from the DOTS rollout. In this way, DOTS can efficiently explore the ground-truth landscape and accelerate the search for global optima. The ratio between top-scored and top-visited samples is a hyperparameter determined to be $5:1$ via grid search (Methods, Supplementary Notes, and Supplementary Table 1).
%that the Top-visited nodes are local 'basins' of the surrogate model. 
%Random samples could increase the input diversity, but not the label diversity.
%Obtaining the labels of these Top-visited nodes would greatly dd
\subsection*{Benchmark problems}
Prior to deploying DOTS to SVL tasks, it is necessary to assess its performance through benchmark studies on high-dimensional, non-convex synthetic problems. To do so, we run extensive benchmark tests on non-convex functions of diverse types (e.g., Ackley, Rastrigin, Rosenbrock, Griewank, Schwefel and Michalewicz functions, see Methods, Supplementary Notes and Supplementary Table 2) and compare the performance of DOTS with other state-of-the-art algorithms. The objective is to find the global minima of these functions with as few data point acquisitions as possible, assuming that we do not know the exact expression of these functions but use them to provide ground truth. Here, we represent and analyze the major results of Ackley, Rastrigin, and Rosenbrock, since these functions are notoriously difficult to find the global optimum. Rastrigin is highly multimodal, having many local minima in the ground-truth landscape. Rosenbrock has a rather long valley with numerous local minima, while Ackley is a mixture type of both functions (Supplementary Fig. 3). These characteristics make them ideal test functions for benchmarking the optimization algorithm under various situations.
%its predictions will inevitably be imperfect.

We evaluate DOTS alongside 11 other state-of-the-art (SOTA) optimization algorithms, encompassing various genres such as exact, heuristic, Bayesian, and tree-based methods. This assessment is conducted under two scenarios: node value assigned by the exact function and by the surrogate model (Supplementary Notes and Supplementary Table 1).   Both scenarios merit investigation. The first scenario may arise when constructing the surrogate model is challenging, while the second scenario is more common, occurring when evaluating the ground truth is prohibitively expensive, necessitating the use of a surrogate model. Therefore, we present the major results from surrogate-model-based optimization of Rastrigin-1000d, Ackley-200d, and Rosenbrock-100d. For each test, we repeat 5 times with different random seeds in Fig. 2a, b and c. 
It is evident that DOTS converges much faster than all baseline algorithms, with some of the baseline algorithms even failing to operate due to memory explosion. Remarkably, DOTS finds the global optimum of Rastrign-1000d within approximately 3000 data points, while other baselines struggle with the immense search space, making little progress. Comprehensive benchmark results of both scenarios can be found in Extended Data Fig. 2 and Supplementary Fig. 4-5. Further, we use convergence ratio to characterize the convergence behavior of the algorithm, which is defined as the frequency of identifying the global optimum within a certain number of iterations, divided by the total number of repetitions. Fig. 2d indicates the convergence ratio on different functions using exact function and surrogate models, it is evident that DOTS can find global optimum of the Rosenbrock-100d and Ackley-200d with very high chance (80$\%$ and 100 $\%$), whereas other baselines are trapped in the local minima and have no chance of finding the global optimum ($\%$). (see Supplementary Table 3-5 for the comprehensive result on the high-dimensional tasks within 1000 dimensions).

To find out the limit of DOTS, We present the convergence result in Table 1, specifying the highest dimension in which DOTS can converge to the global optimum within a limited computing time. (Hardware specification is found in Supplementary Notes). 
The exceptional performance of DOTS is underscored by its ability to converge in up to 2000 dimensions. In contrast, the SOTA methods struggle to converge beyond 100 dimensions. Moreover, while no baselines manage to demonstrate global convergence of Rosenbrock beyond 10 dimensions, DOTS can achieve convergence up to 200 dimensions. For other test functions, DOTS can converge up to 5000 dimensions with exact functions but faces challenges with the surrogate model, indicating difficulties in learning these high-dimensional distributions (i.e., $R^2$ ratio < 0.3, Supplementary Fig. 4 and 41), resulting in inaccurate node values.

%, which is considerably lower than with exact function. This can be attributed to the fact that the surrogate models have difficulty in learning the very high-dimensional distributions, resulting in inaccurate node values.

\begin{table}[H]
   \caption{The max convergence table of the benchmark functions. It shows the maximum dimension at which DOTS and SOTA methods can demonstrate global convergence.}
  \begin{tabular}{ccccc}

     Methods  & Ackley & Rastrigin &  Rosenbrock & Griewank   \\ \hline
    DOTS &  1500 & 2000 &  200 &  500     \\ \hdashline
    SOTA & 100 & 100  & 10  & 60  
             \\ \hline
  \end{tabular}
  \end{table}
%It is noteworthy that in these high-dimensional cases, another tree-search-based baseline (LA-MCTS) always ends up with memory explosion after only a few iterations, further corroborating that using $DUCB$ as an exploration-exploitation trade-off strategy can effectively handle the high-dimensionality.
%with a simulator;  2) coupling with a surrogate model.

Fig. 2e demonstrates the working principle of $DUCB$ inequality.  At each stochastic rollout, DOTS explores the landscape by performing stochastic expansion and computing the $DUCB$ of all leaf nodes, if none of $DUCB$ of leaf nodes is higher than the root node, the inequality rejects the result and the root node remains the same in the next rollout. The rollout will continue till DOTS finds a leaf node with lower $DUCB$ values. However, the stochastic expansion is not able to handle a flat landscape, since the objective is mostly the same and the DOTS could remain at the same local minima for a long time. This problem is effectively solved by local backpropagation.
Figure 2i illustrates how DOTS overcomes local minima in a flat landscape by ascending the 'ladder' formed by local backpropagation. When stuck in the local minima, the number of visitation increases, DOTS updates the $DUCB$ of the root and nearby nodes, creating a local $DUCB$ gradient that allows the algorithm to escape the local optima.  Fig. 2f shows that the top-visit sampling selects both data points with the best scores and the highest visitation number, thereby avoiding the overfitting of the surrogate model. Figure 2j shows the mechanism of adaptive exploration. the exploration weight is no longer a constant, but proportional to $min(y)$, representing the minimum value of the current ground-truth distribution. Consequently, the exploitation and exploration terms in the $DUCB$ formula are dynamically balanced.
%the local backpropagation increases the $DUCB$ of the adjacent nodes 
%In contrast, the global backpropagation would simultaneously update the D-UCB of both near and far nodes, hence the relative $DUCB$ value remains the same.

An ablation study is conducted on Rosenbrock-100d to analyze the impact of individual components of DOTS on overall performance (Fig. 2g, Supplementary Fig. 6 provides additional results). It is evident that local backpropagation serves as the cornerstone of DOTS, while top-visit sampling and adaptive exploration significantly enhance performance and increase the convergence ratio. Without local backpropagation, DOTS essentially becomes a greedy stochastic tree search with the $DUCB$ inequality (where the exploration weight is set to zero), resulting in the poorest performance and an inability to converge. Moreover, without top-visit sampling and adaptive exploration, although DOTS may still achieve convergence sporadically, its performance deteriorates significantly. In such cases, the average best $f(x)$ remains far from the global optimum (around 100), and the corresponding convergence ratio drops to 60$\%$ and 30$\%$, respectively.

Further numerical analysis on the evolution paths of DOTS and its variants supports our observation and provides detailed insights into the individual influence of its components. The evolution paths of DOTS and its ablated variants on Rosenbrock-100d are depicted in both input and ground-truth space (Fig. 2h, k, and l). Figure 2h illustrates the evolution in ground-truth space (with the same initial dataset). It is evident that local backpropagation holds central importance, as DOTS without local backpropagation significantly lags behind all other variants. Moreover, DOTS without adaptive exploration initially evolves faster than the original DOTS in the first 60 rounds but quickly becomes trapped in local minima after 100 rounds. Similarly, DOTS without top-visit sampling keeps pace with the original DOTS but also becomes ensnared in local minima after 100 rounds. Additionally, stochastic expansion and the $DUCB$ inequality also play fundamental roles (refer to Supplementary Fig. 7-10 for the ablation studies).

Figure 2h illustrates the low-dimensional representation of the input distributions from DOTS and its variants using U-MAP (a dimension reduction technique used for visualization) \cite{McInnes2018}, while each plot in Fig. 2i represents the individual evolution path. It can be observed that the original DOTS quickly identifies the region of global optima and eventually converges to it. Meanwhile, DOTS without local backpropagation performs the worst, with none of its data points close to the global optima (Fig. 2i, upper right). DOTS without adaptive exploration (i.e., constant exploration weight) becomes trapped in a local minimum far from the global optima (Fig. 2i, lower left), indicating that the algorithm cannot maintain a balance between exploration and exploitation as the ground-truth distribution shifts due to the addition of new data points. DOTS without top-visit sampling manages to approach the global optima but also fails to escape the local minimum within 150 rounds (additional results can be found in Supplementary Fig. 11).

%mediocre
\begin{figure} [H]
  \centering 
  \includegraphics[width=\textwidth]{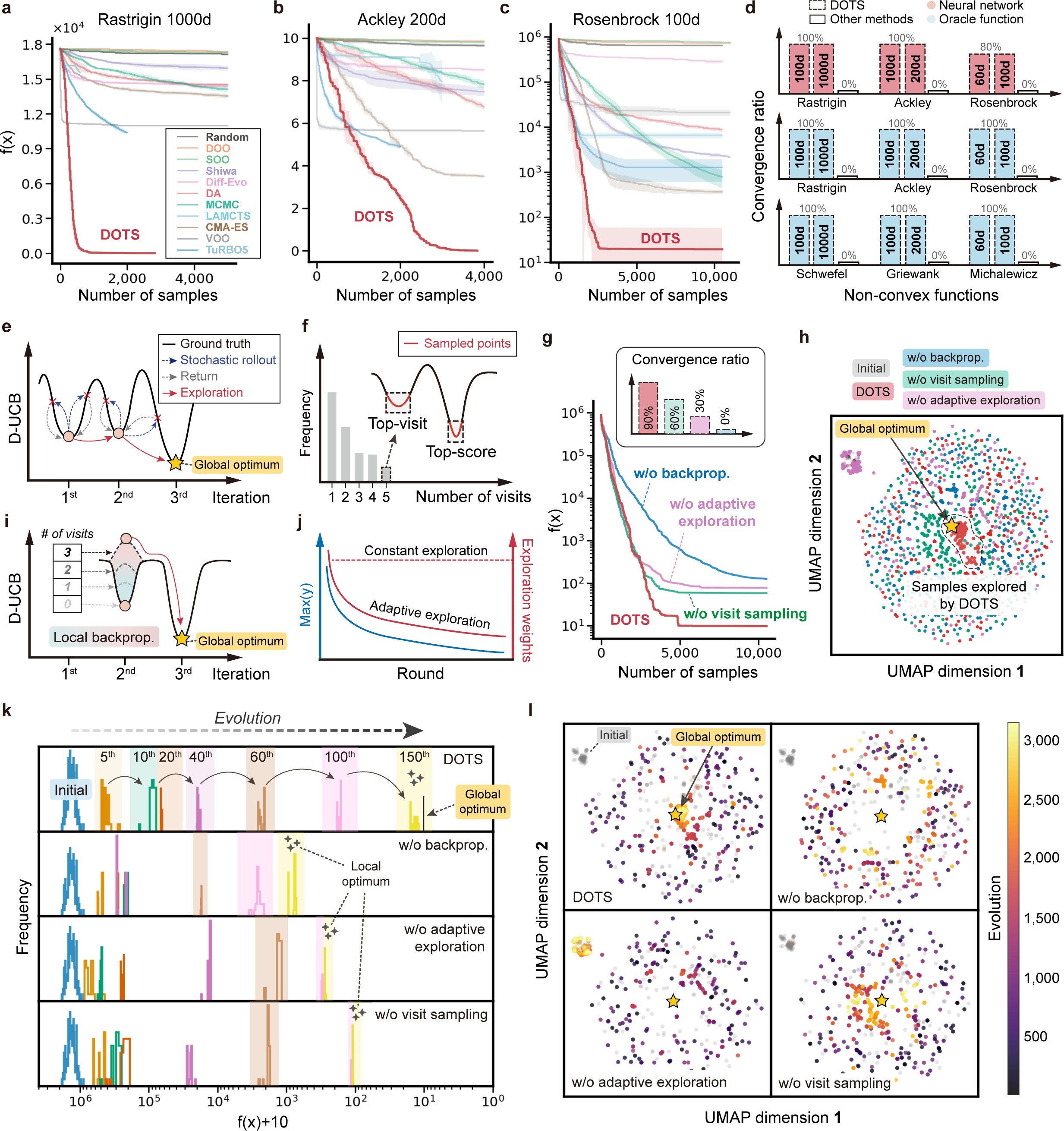}
  \caption{(Caption next page.)}
  \label{fig:2}
\end{figure}
\addtocounter{figure}{-1}
\begin{figure} [H]
    \caption{\textbf{Benchmark study and algorithm analysis (second draft).} \textbf{(a,b,c)} Benchmark studies on Rastrigin 1000d, Ackley 200d, Rosenbrock 100d respectively.  \textbf{(d)} Convergence ratio on various functions of different dimensions. DOTS converges on all synthetic functions with more than $90\%$ ratio, whereas none of the other algorithms can converge at all (0$\%$). \textbf{(e)} DOTS explores a rugged ground-truth landscape using stochastic expansion and $DUCB$ inequality. During stochastic rollout, $DUCB$ of all leaf nodes compare to that of the root, DOTS stays with the root node till leaf node with lower $DUCB$ is found. \textbf{(f)} Top-visit sampling. DOTS iteratively samples data points from other local minima. \textbf{(g)} Ablation study on the individual mechanism. Each has been performed 10 times.  \textbf{(h)} 2-dimensional U-MAP representation of the input distribution of DOTS and its ablated variants on Rosenbrock-100d.
    \textbf{(i)} DOTS overcomes the local minima in a flat landscape by using local backpropagation to create a local $DUCB$ gradient 'ladder'. \textbf{(j)} Adaptive exploration, which keeps the exploration weights at the same level as the overall ground-truth distribution. \textbf{(k)} Evolution path of DOTS in ground-truth space in Rosenbrock-100d. \textbf{(l)} Individual evolution path of DOTS and its ablated variants visualized in separated subplots from \textbf{(h)}.
    }
\end{figure}
%\begin{figure}[H]
    %\centering
    %\includegraphics[width=1\textwidth]%{figs/fig2.4.pdf}

%    \label{fig:2}
%\end{figure}
\subsection*{Self-driving virtual laboratory for complex systems}
SVLs are self-driving laboratories where experiments are virtually carried out by high-fidelity simulators, with the learning models intelligently and iteratively selecting new designs to achieve optimal mechanical, physical, and chemical properties. It is a typical derivative-free optimization problem and of critical importance in many real-world complex systems, particularly when conducting actual experiments is expensive and the design space is extensive.
Achieving optimal design in SVLs demands high efficiency from DFO models, an area where DOTS has demonstrated a clear advantage over other models in benchmark problems. However, beyond dealing with the challenges posed by high dimensionality and non-convexity of the objective function, SVL tasks frequently involve optimizing multiple objectives. Moreover, the search space is often non-linearly constrained due to various external conditions, adding further complexity to the optimization process.
we convert the multi-objective task into a single-objective task by treating the multiple targets as a product of individual targets \cite{WU201990}. Searching for the champion design within a constrained space using DOTS involves adding constraints during stochastic expansion, such as only accepting leaf nodes that satisfy the constraints.
We demonstrate the general applicability of DOTS by applying it to various SVL tasks encountered in four different scientific domains. Specifically, we combine DOTS with four different high-fidelity simulators to iteratively optimize single- or multiple- objectives, i.e., 1) Mechanical property optimization of Architected materials; 2) CCAs design; 3) Cyclic peptide binder design; 4) Optimization of electron ptychography reconstruction. we present the major results of these SVL tasks in Table 2 and the SVL pipelines in Extended Data Fig. 3. 
\subsubsection*{Architected materials design}
Architected materials are one of the most widely adopted engineering materials thanks to their broad applicability and adaptable properties\cite{Xia2022}. Recently, \cite{Peng2023} developed a multi-objective virtual lab pipeline ('GAD') for orthopedic applications. The pipeline combines a learning-based 3D generative model and a high-fidelity simulator to iteratively optimize the architecture to achieve desired elastic modulus (E) and yield strength (Y). 

GAD serves as an ideal benchmark for DOTS (see Supplementary Notes and Supplementary Fig. 12-13 for detailed settings). Here, we adopt the 3*3*3 cubic arrangement of the Gyroid units as the model input and start the SVL task with the same initial labeled dataset adopted by GAD (see Methods).
Figure 3a compares the performance of DOTS with GAD. DOTS identifies superior designs as early as the 3rd round (77MPa) and continues to discover even better designs for the subsequent 5 rounds (82.1MPa). In contrast, although GAD shows rapid improvement in the initial 3 rounds, it fails to make further progress and is eventually outperformed by DOTS after the 5th iteration.
The evolutionary paths (Fig. 3b) demonstrate that GAD tends to concentrate on a few clusters within the design space, whereas DOTS explores a broader region, resulting in continuous improvement and the identification of superior designs.
It is important to note that GAD requires a knowledge-based, large unlabeled dataset to train the generative model for generating new structures, potentially introducing inductive bias. In contrast, DOTS is simpler to implement and free from bias – it relies on stochastic search guided by the machine learning model, without requiring an unlabeled dataset.

The simulated strain-stress curves of the scaffolds depicted in Fig. 3c validate that the DOTS-designed scaffolds exhibit 41.6$\%$ and 9.4$\%$ higher yield strength compared to both the uniformly designed scaffold and the previous GAD scaffold, respectively. The inset of Fig. 3c also presents the density matrix of the DOTS scaffold, revealing a significant concentration of Gyroid units at the tetrahedral points. This non-uniform distribution indicates that DOTS learns to allocate more materials at these crucial junctions, which can greatly enhance structural integrity.
To further investigate the mechanisms underlying their distinct mechanical behaviors, we conduct Finite Element Method (FEM) mechanical analysis on both GAD and DOTS designs. The results depicted in Fig. 3d elucidate the distribution of Von Mises stress and hydrostatic pressure across the DOTS and GAD scaffolds.
GAD tends to distribute mass primarily at the center of the faces of the scaffold, resulting in a commendable 20$\%$ improvement in strength over uniform designs. In contrast, DOTS adopts a different approach by strategically allocating mass preferentially at the vertices and centers of the tetrahedra, achieving impressive strength enhancements compared to both GAD and uniform designs (over 9$\%$ and $41\%$, respectively). From a crystallographic standpoint, the distribution of DOTS tetrahedra forms a close-packed structure, known for its inherent strength and hardness, surpassing the face-centered cubic structure designed by GAD (Additional analysis and results refer to Supplementary Notes, Supplementary Fig.14-15, and Supplementary Table 6.).
%hexagonal
%The results, as depicted in Fig. 2, elucidated the distribution of von Mises stress and hydrostatic pressure across the DOTS and uniform scaffolds (for additional results, refer to Supplementary Fig. 1). Notably, the DOTS scaffold exhibited a mitigated effect of stress concentration compared to its uniform counterpart. Additionally, the hydrostatic pressure plots revealed a higher prevalence of compressed struts within the DOTS scaffolds, as opposed to stretched ones. This phenomenon underscores the efficacy of the DOTS model in strategically allocating materials to optimize stress distribution and bolster structural strength while maintaining a nearly identical mass.

%Therefore, this problem can be treated as a high-dimensional multi-objective optimization problem, presenting an exponential difficulty due to the “curse of dimensionality

\subsubsection*{Compositionally complex alloy design}
%High entropy alloys (HEAs) are an emerging class of materials containing 5 or more major elements. They benefit from high configuration entropy and phase stability and exhibit excellent mechanical or functional properties []. Here, we combine DOTS to design magnetic HEAs that potentially can be used for .. applications??.
%Moreover, superconducting behavior has also been observed in bcc refractory alloys Ta34Nb33Hf8Zr14Ti11\cite{Ko2014}.
CCAs are an emerging class of materials containing 5 or more major elements ($\geq$ 5 at.\%). They have attracted significant and growing interest due to their multi-functional potential, covering mainly mechanical, magnetic, and catalytic properties\cite{nirmal2021}. Another intriguing yet comparatively less-explored aspect of properties pertains to their electronic transport characteristics. In particular, large anomalous Hall conductivity (AHC, $\theta_{xy}$) and anomalous Hall angle (AHA, $\sigma_{xy}/\theta_{xx}$) have been reported in half-Heusler systems such as GdPtBi\cite{Suzuki2016} and TbPtBi\cite{Singha2019}, which hold great promises for sophisticated electronic and spintronic applications. In ferromagnetic CCAs, the chemical disorder is commonly anticipated and adjustable \cite{Suzuki2016,Tanzim2023}. This further motivates us to look for chemically disordered CCAs with higher AHA and AHC.
%In ferromagnetic CCAs, both the ordinary Hall effect (OHE) and anomalous Hall effect (AHE) can be observed. The AHE, 
The expansive chemical space inherent in CCAs poses a considerable challenge. For example, even within a single category of five-component CCAs, there exist approximately 4.6 million combinations with 1 at.$\%$ interval, not to mention the broader array of available elements (i.e., 27 selected elements in this work). Therefore, we build an SVL to systematically search for CCAs that exhibit high anomalous Hall conductivity (AHC) and anomalous Hall angle (AHA).
%in this vast compositional space
%Despite the strong chemical disorder in HEAs, Fermi surface persists in practice \cite{Robarts2020}.  Owing to strong scattering on the intrinsic chemical disorder. For instance, the Cantor alloy CrMnFeCoNi, usually possesses much larger residual resistivity than the binary fcc FeNi or NiCo counterparts\cite{Kud2019}. Actually,
%However, the expansive chemical space inherent in HEAs presents both opportunities and challenges simultaneously. Even within a single category of five-component HEAs, there exist approximately 4.6 million combinations with 1 at.$\%$ interval, not to mention the broader array of available elements (i.e., 27 different elements). %To efficiently explore this vast compositional space,
The SVL for CCAs comprises DOTS and first-principles calculations of anomalous Hall conductivity (AHC) and anomalous Hall angle (AHA) based on the linear response Kubo-Bastin formalism (Methods). The objective is to maximize the product of AHC and AHA while adhering to a constraint on the formation energy threshold, set to 0.02 $\pm$ 0.002 Ry/site. This constraint accounts for the configurational entropy of CCAs, which can counteract the positive formation energy, particularly at around 1300 K (see Methods). We focus on the most common body-centered cubic (bcc) and face-centered cubic (fcc) phases in CCAs \cite{miracle2017critical}. Distinct Fe/(Co+Ni) ratio ranges are selected for the two phases to further rationalize the predicted compositions (see Methods).
%making it a reasonable and simple criterion to ensure the thermodynamic stability of CCAs.  

Fig. 3e demonstrates the target evolution of DOTS and Markov Chain Monte Carlo (MCMC) (technical details are provided in Supplementary Notes and Supplementary Fig. 16-20).
In the initial two iterations, DOTS and MCMC exhibit similar performance on fcc and bcc phases. However, DOTS surpasses MCMC in subsequent iterations, as MCMC likely becomes trapped at local minima, showing no further improvement. The insets depict the distribution of formation energy of predicted alloy compositions, indicating that both algorithms efficiently navigate away from unstable compositions.  Eventually, DOTS achieves higher target values (60 and 82), which are 12.7$\%$ and 28.5$\%$ (fcc and bcc respectively) higher than those of MCMC.
Fig. 3f illustrates the evolution paths of both algorithms using 2D U-MAP representation (see Supplementary Tables 7-10 for compositions of fcc and bcc CCAs predicted by DOTS and MCMC algorithms). 
Regarding bcc CCAs, DOTS focuses on a design space further away from the initial distribution and discovers champion alloy in the 4th iteration, whereas DOTS and MCMC seem to follow similar searching paths in the case of fcc CCAs. However, we note that regardless of crystalline phases, both DOTS and MCMC propose FeCoNiIrAlZn- and FeCoNiIrAlPt-based CCAs in final iterations, suggesting that these elements could play key roles in determining the electronic properties.

To further investigate the mechanism behind the enhanced electronic transport properties of the champion alloy discovered by DOTS, we look into their electronic structures through Bloch spectral functions (BSFs, A$_B$ (E,k)). Fig. 3g showcases the minority-spin Fermi surfaces (FSs) of the champion fcc $\textrm{Fe}_{43.5}\textrm{Co}_{19}\textrm{Ni}_{9.5}\textrm{Ir}_{14.5}\textrm{Al}_{5}\textrm{Zn}_{8.5}$ and bcc $\textrm{Fe}_{63.5}\textrm{Co}_{0.5}\textrm{Ni}_{0.5}\textrm{Ir}_{18.5}\textrm{Al}_{9}\textrm{Zn}_{8}$ CCAs (see also Supplementary Fig. 21-23). By quantifying the degree of smearing using the curve fitting in the minority-spin channel along the selected momentum path \cite{Robarts2020} for the compositions with the best targeted properties in each iteration (Supplementary Fig. 24), we find that the degree of smearing develops in a highly correlated manner with that of AHC*AHA during optimization process (see Fig. 3h). Another noticeable aspect is that the BSFs are more smeared out in CCAs as compared to Fe-Ir binary alloys.
Additionally, FeCoNiIrAlZn-based DOTS CCAs displays higher AHC*AHA than the FeCoNiIrAlPt-based MCMC CCAs. It can be attributed to the fact that the $d$ orbitals of Zn have full band filling (3$d^{10}$) and the $d$-band center is located at a deeper energy level ($\sim -0.5$ Ry) compared to Pt, which as a consequence the band center mismatch among different elements is lager and the disorder scattering is stronger \cite{mu2019uncovering}. DOTS successfully identifies this difference between Zn and Pt, discovers an optimal combination of elements (Supplementary Fig. 24-26).

\subsubsection*{\textit{De novo }cyclic peptide binder design}
%cyclic peptides show better biological activity compared to their linear counterparts due to the conformational rigidity
Cyclic peptides are a class of cyclic compounds that have shown great success as antibiotics and therapeutics in recent years, thanks to their stability, high specificity, and excellent membrane permeability \cite{Vinogradov2019}. The amino acids (AA) in the cyclic peptide are interconnected by amide or other chemically stable bonds between the C– and N–terminal or between the head and tail, representing a high-dimensional and complex sequence design space \cite{Zorzi2017}. Computational design of cyclic peptide binders to target proteins remains challenging \cite{Hosseinzadeh2021,Kosugi2023}. Here, the SVL combines DOTS  with Alphafold2 and Rosetta (protein design software) to design cyclic peptides from scratch with optimized protein-protein interactions (PPIs). 
The optimization target is the product of two Rosetta binding metrics, two widely adopted measures for characterizing the strength of PPIs. (i.e., shape complementarity (SC) and change in Solvent Accessible Surface Area ($dSASA$)\cite{edin2023}, see Methods). The best-fit design is likely to have high $SC$ and $dSASA$. 

%a cyclic peptide complex offset to generate the protein-peptide complex. 
The SVL for cyclic peptide starts with an input consisting of a hotspot region from the target protein, a random AA sequence and a positional encoding which indicates the connectivity. The Alphafold2 takes the input and predicts the corresponding target-peptide complex, from which $SC$ and $dSASA$ can be calculated using Rosseta software \cite{Andrew2011} and feedback to the DOTS. The DOTS iteratively optimizes these metrics by tuning the AA sequence and searching for the binders with higher targets. Here, we directly use the Rosetta-computed metrics as the node value. 
We perform SVL tasks on a selected cyclic peptide-protein dataset, and compared the metrics of resulting designs with that of the gradient descent (GD) and MCMC (Supplementary Table 12) \cite{Kosugi2023}. Fig. 3i shows that starting from the random sequence, the binding metrics of DOTS designs gradually improve and become statistically significantly better than those of GD, MCMC designs, and the native peptides (Mann-Whitney test $P_{GD} =1e-8, P_{MCMC} =2.1e-6, P_{nativel} = 0.048$, alternative = 'greater'). The designed sequences are listed in Supplementary Table 13.
% rather a surrogate model prediction, since it is rather straightforward to obtain the metrics via Rosseta interface analyzer
%A list of all the optimized cyclic peptide-protein complexes and their corresponding sequences and metrics are provided in Supplementary Table 12-13. 

As demonstrated in Supplementary Table 12, the DOTS-designed peptides exhibit higher interface metrics compared to the native binder in multiple cases. Moreover, these designed cyclic peptides adopt a similar binding conformation to the native ones and occupy the same active site, despite having very different sequences. To elucidate the molecular basis behind the differences between the DOTS designs and the native binder (Supplementary Table 13), we conduct an interaction map analysis of one example, focusing on the casein kinase 2 (CK2) and the DOTS-designed cyclic peptide (referred to herein as the DOTS peptide).
%in complex with either the native binder (PDBID: 4ib5) 
%In the CK2$\alpha $1-335/Pc cocrystals (PDBID: 4ib5), Pc as a native cyclic peptide binder binds$ CK2\alpha$ 1-335 subunit A via the known CK2 $\beta$-binding site at the outer surface of the N-terminal $\beta$-sheet. The binding site features a distinctive architecture where a hydrophobic cavity is situated adjacent to a solvent-accessible surface. This cavity is encircled by key residues, including Y37, V65, V99, and A108. The Pc ligand plunges deeply into the hydrophobic cavity with a phenyl group of residue F2. Within the native cyclic peptide binder Pc, the hydrophilic amino acids Y13, K3, and H5 are pivotal, engaging in interactions with crucial CK2 residues including Q34, E50, and L39.

The DOTS peptide exhibits enhanced interactions compared to the native peptide (DOTS: $SC$ = 0.79, $dSASA$ = 1228 $\AA^2$ vs. Native: $SC$ = 0.63, $dSASA$ = 1046 $\AA^2$). This enhancement can be attributed to the fact that DOTS introduces larger hydrophobic groups near the interface and increased hydrophilic contacts (as compared to the native binder, as shown in Extended Data Fig. 4). The Alphafold2 predicted complex, depicted in Fig. 3j, illustrates a hydrogen-bonding network. In addition to the interactions between the peptide and residues Q34, E50, and L39, which mirror those observed with the native binder, the imidazole groups of residues H1 and H11 establish additional salt bridges with the acidic side chains of D101 and E50, respectively. Moreover, the indole ring of the W2 residue fits into a hydrophobic cavity, similar to the native binder, establishing interactions with adjacent hydrophobic residues, including Y37, V65, V99, and A108 (refer to Fig. 3k). This engagement facilitates the anchoring of the peptide to the protein surface, thereby stabilizing the peptide-protein complex. Additionally, the DOTS peptide incorporates the hydrophobic amino acid F4 close to the receptor protein (L39 and F52), enhancing their shape compatibility. Hence, despite having different sequences compared to the native peptide, DOTS manages to maintain a similar conformation to the native peptide while achieving good stability and high binding metrics by striking a balance between hydrophobic and hydrophilic interactions. Similar patterns for other DOTS designs are observed, and detailed comparisons are provided in Supplementary Fig. 27-36.
%the carbonyl group of residue Cys13 establishes hydrogen bonds with residue Gln34.
%Moreover, a similar interaction is observed between the imidazole group of H11 and the acidic side chain of E50.
%while retaining most hydrophilic and hydrophobic interactions,
\subsubsection*{Electron ptychography reconstruction optimization}
Scanning transmission electron microscopes (STEM) with aberration correctors have been capable of characterizing nanostructures at a sub-angstrom resolution\cite{Peter2011}. However, atomic-resolution imaging of nanostructures by STEM is often hindered by multiple electron scattering in samples thicker than a monolayer\cite{Cowley1957}, as well as by beam-induced damage in sensitive materials\cite{Song2019}. Electron ptychography, a phase-contrast imaging technique, has emerged as a promising solution to overcome these challenges and achieve sub-angstrom resolution and three-dimensional depth sectioning in samples thicker than 20 nm \cite{Zhen2021}. The goal of ptychographic reconstruction is to quantify the phase of the transmission function within the atomic lattice. The quality of this reconstruction relies on a careful selection of various reconstruction parameters, including physical, optimization, and experimental parameters, which collectively affect the quality and accuracy of the retrieved transmission function. The parameter space can be vast and complex, and the optimal choice depends on the specific dataset and measurement conditions. Currently, the parameter selection is mainly based on expert knowledge and trial-and-error, which limits the efficiency and applicability of electron ptychography. 
Here, the SVL task involves finding the optimal reconstruction parameters to retrieve the underlying transmission function. This task is equivalent to solving a non-convex problem in an 8-dimensional parameter space (for technical details, refer to Supplementary Table 14).
The objective is to iteratively minimize the normalized mean square error (NMSE) between the measured and modeled diffraction patterns (Extended Data Fig. 5).

%This is achieved by reducing the normalized mean square error (NMSE) initial and revised wave functions, which can be characterized by 7-dimension parameter spaces (detailed parameters and their bounds are listed in Table S1).  Electron ptychography 
 %This requires solving a non-convex problem in a 7D parameter space in our case (detailed parameters and their bounds are listed in Table S1).
%The objective function is the normalized mean square error (NMSE) between the initial and revised wave functions, $\phi$ and $\phi'$, respectively. The initial wave function $\phi$ is the product of the initial guesses of the probe and object functions, P and O. The diffraction pattern of $\phi$ is modeled by a Fourier transform, and its amplitude is replaced by the square root of the measured diffraction intensity before inverse Fourier transforming back to the image plane, yielding $\phi'$. The difference  $\Delta \phi$ between$\phi'$ and $\phi$ is used to update both P and O. By minimizing the NMSE, O converges to the true transmission function, as shown in Fig. 3 (column 2).

We apply DOTS to this SVL task and compare it with the built-in BO and TuRBO5 algorithms. To this end, we simulate a 4D dataset of 18-nm-thick silicon along the [110] direction with Poisson noise. Fig. 3l summarizes the benchmark result (Supplementary Table 15). DOTS achieves a final NMSE of 0.291, outperforming the baseline BO in almost every iteration. Although TuRBO5 shows a comparable result with DOTS and reaches an NMSE of 0.293, its parameter set appears to be a local optimum that disagrees strongly with the ground-truth parameters on the electron probe (i.e. beam energy 238 kV and defocus 129 \AA, versus 200 kV and 100 \AA, respectively; more details are listed in Supplementary Table 15).
The reconstruction quality of different methods is assessed by the correlation index, which compares the phase of the simulated and reconstructed transmission functions (Methods). Again, DOTS achieves the highest score of 0.958, even surpassing the quality of expert's choice based on the ground truth simulation and the information of the inter-planar spacing of silicon (Fig. 3m). Moreover, Fig. 3n shows the 1D distance profiles of the phase for different silicon positions by the simulated and DOTS-reconstructed transmission functions. The relative entropy, ranging from 0 to 1 where 0 (1) indicates identical (distinct) distributions, between the two distributions is 0.029. This result further confirms that parameters selected by DOTS lead to near-perfect reconstructions. The slight blurring at the edges of the silicon atom positions could be attributed to the effect of Poisson noise in the data.
\begin{table}[h]
   \caption{The overall performance of the SVL tasks. The increase is defined as the relative increase ratio achieved by DOTS.}
  \begin{tabular}{lccc}

   Domain & Optimization target(s) & Search space & Increase ($\%$) \\ \hline
    Architected materials  & Mechanical properties  & $8^{27}$ &  +9.4               \\ \hdashline
    CCAs  & Electronic transport properties  & $100^{27}$  & +28.5                \\ \hdashline
     Cyclic peptide binder & PPIs   & $8^{20} - 20^{20}$ & +33.7                  \\ \hdashline
    Electron ptycho. recon. & Resolution  & $20^{8}$ & +2.0                  \\ \hline
  \end{tabular}
  \label{tab:2}
  \end{table}
\begin{figure} [H]
  \centering 
  \includegraphics[width=\textwidth]{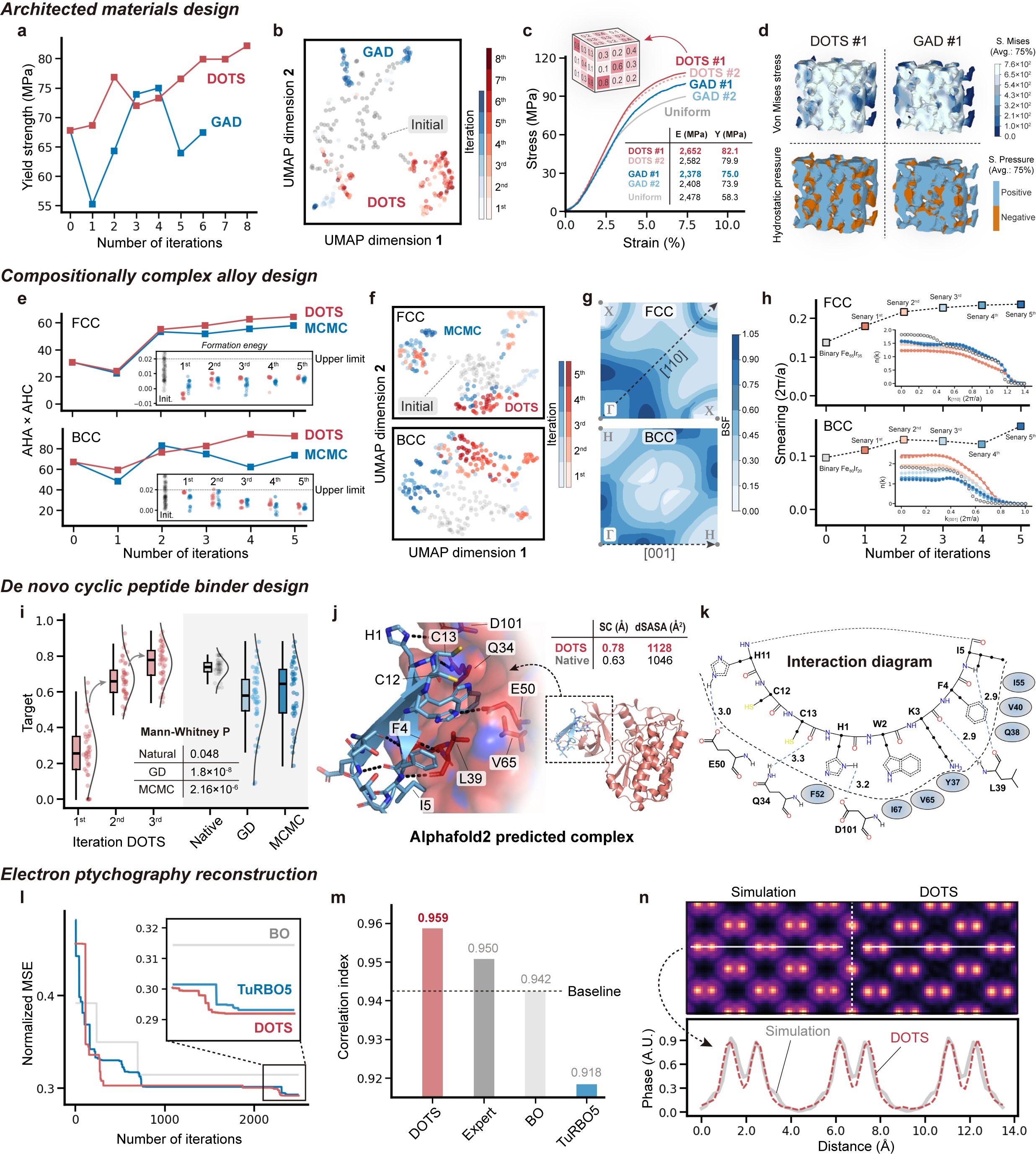}
  \caption{(Caption next page.)}
  \label{fig:3}
\end{figure}
\addtocounter{figure}{-1}
\begin{figure} [H]
\caption{\textbf{Self-driving virtual laboratories for complex systems.} \textbf{(a)} Optimizing the mechanical properties of architected materials by DOTS and GAD (baseline). \textbf{(b)} U-MAP 2D representation of input distribution from both methods.\textbf{(c)} Stimulated strain-stress curve of both methods. The inlet shows the density matrix. \textbf{(d)} Von Mises stress and hydrostatic pressure across the DOTS and GAD scaffolds. \textbf{(e)} Optimizing the electronic properties of  CCAs by DOTS and MCMC (baseline). \textbf{(f)} U-MAP 2D representation of input distribution from both methods. \textbf{(g)} Fermi surface of DOTS-designed CCAs (bcc and fcc). \textbf{(h)} Fermi surface smearing. The smearings increase as the iteration increases. Inlets show the curves along a selected momentum path on the Fermi surface, a quantitative measure for describing the smearing. \textbf{(i)} Optimizing the PPIs using  DOTS and other two methods. \textbf{(j)} Alphafold2 predicted complex, where the cyclic peptide is designed by DOTS. \textbf{(k)} Interaction diagram of DOTS peptide with the target protein. \textbf{(l)} NMSE loss history of DOTS, BO and TuRBO5 \textbf{(n)} Correlation index of between the phase reconstructions (DOTS, expert, BO and TuRBO5) and ground-truth. \textbf{(m)} The simulated and DOTS-reconstructed transmission functions, where the 1D distance profiles are shown. }
\end{figure}

\section*{Discussions}
We have introduced a tree search method for optimization that addresses high-dimensional, non-linear problems where the structure of the underlying objective function is entirely unknown. This method targets a significant category of optimization tasks for which computationally tractable and efficient solutions have been challenging to achieve. DOTS demonstrates superiority over existing state-of-the-art algorithms for various high-dimensional non-linear tasks by achieving convergence beyond 2000 dimensions and maintaining 100$\%$ convergence ratios in various benchmark problems at 1000 dimensions. In contrast, most existing algorithms struggle to find global optima beyond 20 dimensions.
Looking ahead, the current bottleneck lies in the expressive power of the surrogate model and the available computer memory, rather than in the capacity of the DOTS algorithm itself. Currently, DOTS is being executed on a desktop computer setup. There is potential for DOTS to further push the boundaries of dimensionality by employing more sophisticated surrogate models and leveraging larger computing resources. This may enable DOTS to successfully address problems beyond 2000 dimensions.
% It is worth reiterating that many learning algorithms only operate at a rather low dimension (below 100 dimension), whereas DOTS has effectively solved the benchmark problems and extended them to a new dimension, finding the global optima of synthetic function even beyond 2000 dimension within a rather reasonable computational cost. 

Furthermore, DOTS has showcased its efficacy in executing constrained multi-objective optimization across a range of virtual laboratory tasks, encompassing structure, alloy, protein design, and electron microscopy. This includes achieving desired objectives while adhering to external constraints. Looking ahead, we envision ample opportunities to apply our method across various quantitative sciences where optimization is crucial, as our approach can seamlessly integrate with any simulator or experimental setup. One particularly promising avenue for future application is to merge our method with a robotic setup to facilitate real-world automated experimental design, thereby enabling accelerated materials discovery or synthesis.
Indeed, another potential application lies in financial optimization, where the objective is to allocate resources optimally to maximize returns or achieve specific financial goals. We foresee that our algorithms will soon become standard practice and will be integrated with virtual or experimental setups from other disciplines to address high-dimensional and non-linear optimization tasks that were previously considered beyond reach. This interdisciplinary approach holds great promise for unlocking new solutions and advancing various fields of research and practice.

\section*{Methods}

\subsection*{Machine learning surrogate model}
We implement a 1D convolutional neural network for the surrogate model. It consists of 1D convolutional layers and is followed by pooling, dropout, and normalization layers to prevent overfitting. The network parameters are optimized using Adam Optimizer, and the loss function is the mean-squared error or mean absolute percentage error. 200 initial data is randomly generated, both for benchmark and VSL tasks. The algorithm is implemented using TensorFlow 2.0. More detailed parameters are found in Supplementary Notes and Supplementary Fig. 37.

\subsection*{Synthetic functions}
The synthetic functions are designed for evaluating and analyzing the computational optimization approaches. In total, six of them are selected based on their physical properties and shapes. Results of Ackley, Rosenbrock, and Rastrigin functions are presented in the main text since they are popular and relevant results are available in much literature. We also test three other synthetic functions (Griewank, Schwefel, Michalewicz), and the results are presented in Supplementary Fig. 5.
The Ackley function can be written as:
\begin{equation}
   f(x) = -a \cdot exp(-b\sqrt{\frac{1}{d}\sum_{i=1}^{d}x_i^2}-exp (\frac{1}{d}\sum_{i=1}^{d}cos(cx_i))+a+exp(1)
\end{equation}
where $a$ = 20, $b$ = 0.2, $c$ = $2\cdot \pi $, and $d$ is the dimension.

The Rosenbrock function can be written as:
\begin{equation}
   f(x) = \sum_{i=1}^{d-1}[100(x_{i+1}-x_{i}^2)^2+(x_i-1)^2]  
\end{equation}

The Rastrigin function can be written as:
\begin{equation}
   f(x) = 10d +  \sum_{i=1}^{d-1}[x_i^2 - 10cos(2\pi x_i)]
\end{equation}
The three functions are evaluated on the hypercube $x_i \in$ [-5, 5], for all $i$ = 1, …, $d$ with a discrete search space of a step size of 0.1, we also show that different step sizes (within a certain range) do not affect the general behavior of the algorithm (see Supplementary Fig. 1). 
20 samples per round when using neural networks as surrogate models. More details and results are found in Supplementary Notes.
%\subsection*{Traveling salesman}
%\textbf{Feature engineering:}\\
%\textbf{Optimization target:}\\

\subsection*{Architected materials}

\textbf{Feature engineering:} In this study, the objective for architected materials optimization is a Gyroid triply periodic minimal surface (TPMS) structure, which naturally occurs in butterfly wings and is renowned for its exceptional biological characteristics and mechanical performance. The Gyroid scaffold to be optimized comprises 27 subunits with a dimension of 2*2*2 mm, allowing for tuning its geometry features and mechanical properties by adjusting each subunit's density. The density of each subunit can take discrete values from 10 to 80$\%$, with an increment of 10$\%$. The base material of the scaffold is Ti6Al4V alloy. 3D convolutional neural networks are employed to accurately and rapidly assess the impact of the adjustments of the subunit's density on the scaffold's performance. Details about structure generation are presented in ref\cite{Peng2023}.\\
\textbf{Optimization Target:} To mechanically stimulate bone reconstruction in bone defects, it is well-recognized that the elastic modulus of bone grafts should be equivalent to that of the replaced bone, which ranges from 0.03-3 GPa for cancellous bone and 3-30 GPa for cortical bone while there are specific modulus demands for different anatomical locations \cite{wang2016topological}. Moreover, it requires the optimization of load-bearing capacity to prevent damage during implantation. Here, we establish the modulus requirement for the implanted site at 2.5 GPa. Consequently, the optimization target is to maximize the yield strength of the scaffold while ensuring the elastic modulus remains within a specified range (2500  $\pm$ 200 MPa).\\
\textbf{Finite element simulation:} 
Finite element (FE) simulations of the compressive stress-strain curves of scaffolds are conducted using ABAQUS 2018. The FE simulations utilize the same rigid-cylinder and deformable-implant-structure model. The material property is set to be homogeneous with a Poisson's ratio of 0.25, more details in the calibration protocol were developed in the ref\cite{Peng2023}. Ductile damage is employed to simulate plastic deformation up to the failure stage, with a fracture strain set at 0.03. The effects of triaxiality deviation and strain rate are disregarded. Displacement and force are extracted during post-processing and subsequently converted to strain and stress, respectively. FE simulation agrees well with the experiment compression curves (Supplementary Fig. 38). \\
\textbf{Machine learning model:} Initial dataset (100 density matrices) is consistent with our previous work \cite{Peng2023} and the corresponding elastic modulus and yield strength are calculated by FE simulations. 3D convolutional neural networks are employed to predict the elastic modulus and yield strength of the scaffolds with varying density matrices. The model architecture comprises an input layer, convolutional layers, fully connected layers, and an output layer (refer to Supplementary Notes and  Supplementary Fig. 12 for detailed parameters). In the input layer, the scaffold structure is voxelized into 60*60*60 pixels, where each pixel denotes either the solid phase (1) or void phase (0) within the scaffold. The convolutional layers are designed with a series of 3D convolution kernels to extract high-dimension information about the scaffold, while the output layer delivers the final prediction.

\subsection*{Compositionally complex alloys}
\textbf{Feature engineering:} We adopt 27 elements: Fe, Co, Ni, Ta, Al, Ti, Nb, Ge, Au, Pd, Zn, Ga, Mo, Cu, Pt, Sn, Cr, Mn, Mg, Si, Ru, Rh, Hf, W, Re, Ir, and Bi, to design 6-element CCAs with either bcc or fcc structures. For Fe, Co, and Ni, the atomic ratio ranges from 0 to 100 at.$\%$, while for other elements, it ranges from 0 to 40 at.$\%$, with 0.5 at.$\%$ interval. Additionally, the total atomic percentage of Fe, Co, and Ni is designed to fall between 60 at.$\%$ to 80 at.$\%$. For CCAs with a bcc grain structure, the Fe / (Co + Ni) ratio is required to be greater than or equal to 1.5, whereas for fcc structures, it is required to be less than or equal to 1.5.\\
\textbf{Optimization Target:} The optimization target is to maximize the following target:
\begin{equation}
   Target=AHC \cdot AHA
\end{equation}
While keeping the formation energy under the upper limit of 0.02.\\
\textbf{Density functional calculation:} The transport properties are described by the conductivity tensor $\sigma_{\nu \mu}$ ($\nu,\mu=x,y,z$). The AHC ($\sigma_{xy}$) and AHA ($\sigma_{xy}/\sigma_{xx}$) are determined in the frame of Kubo-Bastin (KB) linear response formalism within relativistic multiple-scattering Korringa-Kohn-Rostoker (KKR) Green’s function (GF) method\cite{Hubert2015}, which has been implemented in the MUNICH SPR-KKR package\cite{Ebert2011,Jan2014}. The KB formalism includes both the Fermi-surface and Fermi-sea contributions to equal footing, in which the Fermi-surface term contains only contribution from states at the Fermi energy ($E_F$) while the Fermi-sea term involves all the occupied states (with energy $E$) below the Fermi energy, i.e.,
\begin{align}
    \sigma_{\mu \upsilon} &= \sigma_{\mu \upsilon}^I + \sigma_{\mu \upsilon}^{II}\\
     \sigma_{\mu \upsilon}^I &= \frac{\hbar}{2 \pi \Omega}Tr\langle\hat{j}_{\mu}(\hat{G}^+ - \hat{G}^-)\hat{j}_v\hat{G}^- - \hat{j}_{\mu}\hat{G}^+\hat{j}_v(\hat{G}^+ - \hat{G}^-) \rangle\\
     \sigma_{\mu \upsilon}^{II} &= \frac{\hbar}{2 \pi \Omega}\int_{-\infty}^{E_F}Tr\langle\hat{j}_{\mu}\hat{G}^+\hat{j}_v \frac{d\hat{G}^+}{dE} - \hat{j}_{\mu}\frac{d\hat{G}^+}{dE} \hat{j}_{v}\hat{G}^+  - (\hat{j}_{\mu}\hat{G}^-\hat{j}_v \frac{d\hat{G}^-}{dE} - \hat{j}_{\mu}\frac{d\hat{G}^-}{dE} \hat{j}_{v}\hat{G}^-) \rangle
\end{align}
The electric current operator is given by$\hat{j}_{\mu(v)}=-|e|c\alpha$, with $e>0$ being the elementary charge. $\hat{G}^+$ and $\hat{G}^-$ denote the retarded and advanced GFs, respectively. The representation of the GFs for the first-principles treatment of Eqs. 10 and 11 lead to a product expression containing matrix elements of the current operators with the basis functions and k-space integrals over scattering path operators. In this averaging procedure, the chemical disorder and vertex corrections are treated by means of coherent potential approximation (CPA)\cite{Butler1985}. For both Fermi surface and surface terms, the conductivity tensor partitions into an on-site term $\sigma^0$ involving regular and irregular solutions and an off-site term $\sigma^1$  containing only regular solutions. This formalism has been validated to provide consistent residual and anomalous Hall resistivities with experiments \cite{Kud2019,Hubert2015}, and more detail can be found in Supplementary Notes.\\
\textbf{Machine learning model:} Initial 200 CCAs are randomly generated following the previously described design rules, and their corresponding AHA, AHC, and formation energy are calculated by DFT. For CCAs with bcc grain structures, 154 configurations ultimately converge in the DFT calculations, whereas for fcc structures, there are 178. We train 1D convolutional neural networks to predict the AHA, AHC, and formation energy of the CCAs. The model architecture includes an input layer, convolutional layers, fully connected layers, and an output layer (see Supplementary Notes and  Supplementary Fig. 16 for detailed parameters).

\subsection*{Cyclic peptide binder}
\textbf{Feature engineering:}
We represent the cyclic peptide as a sequence of integers that range from 0 to 19, with each number corresponding to a distinct type of canonical amino acid. The leaf node within the DOTS framework is obtained through stochastic expansion.\\
\textbf{Optimization Target:} 
The optimization target of cyclic peptide binder is defined as follows:
\begin{equation}
   Target=SC \cdot dSASA / 100
\end{equation}
where $SC$ stands for Shape Complementarity, and $dSASA$ represents the change in Solvent Accessible Surface Area before and after interface formation. The SC value ranges from 0 to 1, referring to how well the surfaces of two proteins fit geometrically together at their interface; $dSASA$ measures the size of the interface (in units of $\AA^2$). Both metrics are essential for assessing the quality of the interface. Therefore, we multiply these two metrics to formulate a multi-objective optimization problem, which is used to evaluate the performance of DOTS.\\
\textbf{Dataset:} 
Fourteen unique protein and canonical cyclic peptide complexes aresourced from the Protein Data Bank (PDB), with peptide lengths ranging from 7 to 14 amino acids. We perform 3 different optimization tasks using DOTS, GD, and MCMC. The tasks start from a random initial sequence. The structure with the highest target value is selected as the best structure. For each task, we perform three independent tests. \\
\textbf{Alphafold2 Settings:}
The structure of protein and cyclic peptide binder complex is predicted by Alphafold2-multimer implemented in ColabDesign. A modified offset matrix for the relative positional encoding of a target protein and cyclic peptide complex is adapted to give the structure with high accuracy \cite{Kosugi2023}. For designing a cyclic peptide binder, the binder hallucination protocol is utilized for both GD and MCMC methods. In this study, we maintain the length of the cyclic binder and the interaction site hotspots consistent with those found in nature. For GD, the method $'design\_pssm\_semigreedy()'$ is employed, setting $soft\_iter$ to 120 and $hard\_iter$ to 32. The loss function is a weighted sum of pLDDT (predicted Local Distance Difference Test) and interface contact loss, with other parameters left at their default settings. For the MCMC method, a total of 1000 steps are executed to find the sequence achieving the highest pLDDT. More detail can be found in Supplementary Notes.\\
\textbf{Rosetta Interface Analyzer:}
The $SC$ and $dSASA$ values for the predicted structure of the protein and cyclic peptide complex are computed using the Rosetta Interface Analyzer. Initially, the Rosetta minimize protocol is applied to obtain the structure with minimum energy proximal to the initial conformation. To ensure that cyclic peptides within the complex retain their cyclic nature and do not become linear, the options $'-use\_truncated\_termini'$ and $'-relax:bb\_move false'$ are employed. Subsequently, the minimized complex serves as the input for the interface analyzer.\\
%\subsection*{Sampling}
\subsection*{Electron ptychography}
\textbf{Feature engineering:} The feature vector consists of 8 variables: beam energy, defocus, maximum number of iterations, number of iterations with identical slices, probe-forming semi-angle, update step size, slice thickness and number of slices. Detailed values and their bounds are listed in Supplementary Table 14. \\
\textbf{Optimization Target:} The objective function NMSE is calculated between the positive square-root of the measured diffraction pattern $I_{M}$ and the modulus of the Fourier-transformed simulated exit-wave $\Psi$, which can be formulated as: 
\begin{equation}
   \frac{1}{N}\sum_{i}^{N}\left | \sqrt{I_{M(i)}(\mathbf{u})} - \left | \mathcal{F} [\Psi_{i}(\mathbf{r})] \right | \right |^{2}
\end{equation}
where $\mathbf{r}$ and $\mathbf{u}$ denote the real- and reciprocal-space coordinate vectors, respectively, and $N$ is the total number of the measured diffraction patterns.\\
\textbf{Correlation index:} The degree of matching for a given template $T$  by intensity function $P$ is characterized by a correlation index, which can be defined by the following relation:
\begin{equation}
 \frac{\sum_{i=1}^{m} P(x_i,y_i)T(x_i,y_i) }{\sqrt{\sum_{i=1}^{m} P^2(x_i,y_i)}\sqrt{\sum_{i=1}^{m} T^2(x_i,y_i)}}
\end{equation}
where $(x_i, y_i)$ is the coordinate of pixel $i$. \\
%All methods start with 20 points and run for 2,500 iterations.
\textbf{Dataset simulation:} abTEM\cite{MADSEN2021}, an open-source package, is used for the simulation of a transmission electron microscopy experiment. For this case study, we simulated a 4D dataset of 18-nm-thick silicon along the [110] direction with Poisson noise.\\
\textbf{Ptychographic reconstruction:} The analysis is performed using py4DSTEM\cite{Savi2021}, a versatile open-source package for different modes of STEM data analysis. See Supplementary Notes and  Supplementary Fig. 39 for more details about the reconstruction process.

\hfill \break
%\textbf{Acknowledgements:}

\hfill \break
\textbf{Author contributions:}
Y.W. conceived the idea; Y.W. and B.P. developed the theory and methods. B.P. and Y.W. implemented the algorithms. B.P., Y.W., R.X, P.Y.T., Y.Q., Y.C and S.B. carried out the numerical studies and analysis; R.X. performed the DFT calculations; Y.Q. performed the FEM analysis; Y.C. built the cyclic peptide design pipeline; P.Y.T. developed the electron ptychography simulation pipeline; P.Y.T., B.P. and Y.W. produced the final figures; Y.W. and B.P. wrote the original draft; All authors contributed to data analysis, discussion and manuscript.

\hfill \break
\textbf{Competing interests:} Authors declare that they have no competing interests.
\hfill \break
\textbf{Code availability:} Code and data for DOTS are available at:\\\href{https://github.com/Bop2000/DOTS/}
{https://github.com/Bop2000/DOTS/}
\hfill \break
\textbf{Supplementary Information:}\\
%Section S1 to S5 \\
%Figure S1 to S23 \\
%Table S1 to S5 

\bibliography{main}

\bibliographystyle{Science}

\end{document}